\documentclass{article}
\usepackage{spconf,amsmath,graphicx}
\usepackage{multirow}
\usepackage[utf8]{inputenc}
\usepackage[dvipsnames]{xcolor} 
\usepackage{amsmath,amsfonts,amssymb,amsbsy, amsthm,ae,aecompl}
\usepackage[toc,page]{appendix}
\usepackage{algpseudocode}
\usepackage{epsfig}
\usepackage{comment}
\usepackage{amsmath}
\usepackage{hyperref}
\usepackage{bm}
\usepackage{tikz}
\usetikzlibrary{arrows}
\usepackage[dvipsnames]{xcolor} 
\usepackage[caption=false,font=footnotesize]{subfig}
\usepackage{tikz,pgfplots}
\usepackage{algpseudocode}

\newtheorem{corollary}{Corollary}
\newtheorem{lemma}{Lemma}

\title{Improving Classification Accuracy with Graph Filtering}
\name{M. Hamidouche$^\star$, C. Lassance$^\star$, Y. Hu$^\dagger$, L. Drumetz$^\star$, B. Pasdeloup$^\star$, V. Gripon$^\star$}

\address{
\begin{minipage}{0.5\textwidth}
\begin{center}
$^\star$
IMT Atlantique, Lab-STICC\\
Brest, France
\end{center}
\end{minipage}
\hfill
\begin{minipage}{0.5\textwidth}
\begin{center}
$^\dagger$
Orange Labs\\
Rennes, France
\end{center}
\end{minipage}}
\begin{document}
%
\maketitle
\begin{abstract}

In machine learning, classifiers are typically susceptible to noise in the training data. In this work, we aim at reducing intra-class noise with the help of graph filtering to improve the classification performance. Considered graphs are obtained by connecting samples of the training set that belong to a same class depending on the similarity of their representation in a latent space. 
We show that the proposed graph filtering methodology has the effect of asymptotically reducing intra-class variance, while maintaining the mean. 
While our approach applies to all classification problems in general, it is particularly useful in few-shot settings, where intra-class noise can have a huge impact due to the small sample selection. Using standardized benchmarks in the field of vision, we empirically demonstrate the ability of the proposed method to slightly improve state-of-the-art results in both cases of few-shot and standard classification.

\end{abstract}

\section{ Introduction}
\label{sec:intro}

Deep learning has experienced tremendous growth in recent years and is considered today of major relevance for computer vision tasks such as object detection \cite{zhao2019object} or image classification \cite{lu2007survey}. The achieved top performance is due to the sophistication of the training algorithms and of the trained models, the growing computing capability of the machines, and the availability of large datasets. 

As depicted in Figure~\ref{fig:workflow}, a deep learning model can be interpreted as the concatenation of: a) a \emph{feature extractor}, that generates high-level features from a processed input, and b) a \emph{classifier}, that maps these features to a particular class. Both blocks are generally trained jointly, such that this vision of a deep learning model in two blocks is only useful for interpretation purposes.


Similarly to other classifiers, deep learning methods are susceptible to noise in the training data which can lead to reduced generalization abilities and lack of robustness. Trying to reduce this noise in the raw input domain of data is complicated and could result in overall poorer performance if performed too aggressively. 

On the contrary, the latent representation of input samples, provided by the feature extractor, is a better representation for denoising, as it can be viewed as a decomposition of the raw data into more abstract features. For a fixed class, an efficient filtering procedure would therefore nullify individual specificities of labeled samples in that latent space, to improve the representation of the class more generally.

From a signal processing perspective, a sample's own specificities can be seen as high frequencies in an underlying space modeling its class. The framework of Graph Signal Processing (GSP)~\cite{shuman2013emerging} provides the exact tools we need to model such a domain through a graph representation, where vertices correspond to labeled samples of a given class, and features are seen as graph signals.  In this work, we aim at constructing one graph for each class, using these graphs to filter not the raw input images of the considered class but instead high-level features typically obtained at a deep layer of a trained deep neural network, corresponding to the output of the previously introduced \emph{feature extractor}.


We conduct this analysis to improve the accuracy of the considered classifier. We expect highest gains in the case of few-shot, where only few labeled samples are available for each class. It is worth to mention that in contrast with standard classification, in the few-shot setting the feature vectors that are extracted by a deep neural network are trained on a large dataset (base classes,  $\mathcal{D}_{\mathrm{base}}$). These features are then specialized to a smaller and distinct dataset (novel classes,  $\mathcal{D}_{\mathrm{novel}}$), generally through \emph{transfer-learning} methods~\cite{torrey2010transfer,rohrbach2013transfer}.

The outline of the paper is as follows. In Section~\ref{sec:methodology} we introduce the proposed methodology. In Section~\ref{sec:theory} we show on simplified settings that the filtering operation asymptotically reduces intra-class variance and preserves the mean. 
In Section~\ref{sec:numerical-result} we perform experiments on standardized benchmarks, and show that the proposed methodology can improve the classification accuracy even in the case of state-of-the-art solutions. We consider both few-shot and classical settings. Finally, Section~\ref{sec:conclusion} is a conclusion.

Throughout this article, we use the following notations. Real values and functions are given in italic (e.g., $m$, $h$). Vectors are denoted in bold lowercase (e.g., $\mathbf{x}$), with $i$-th entry as $\mathbf{x}_i$, letters in bold uppercase (e.g., $\mathbf{W}$) denote matrices, with $i$-th row as $\mathbf{W}_{i,:}$, $j$-th column as $\mathbf{W}_{:,j}$, submatrix of $k$ first columns as $\mathbf{W}_{:,:k}$, and entry at intersection of row $i$ and column $j$ as $\mathbf{W}_{i,j}$. 
Finally, $|\cdot|$ is the cardinal of a set, and $\mathrm{diag}(\cdot)$ builds a diagonal matrix from a given vector.

\section{Proposed Method}
\label{sec:methodology}

\begin{figure}
    \centering
    \begin{tikzpicture}[scale=0.9, thick]
    \node[inner sep=0pt](0) at (0,0) {\footnotesize{\begin{tabular}{c}input\\$\mathbf{X}$\end{tabular}}};
    \node[inner sep = 0pt, draw, rectangle, minimum height=1.3cm](1) at (1.75,0) {\footnotesize{\begin{tabular}{c}feature\\extractor\\$\phi$\end{tabular}}};
    \node[inner sep = 0pt](2) at (4,0) {\footnotesize{\begin{tabular}{c}feature\\tensor\\$\mathbf{F} = \phi (\mathbf{X})$\end{tabular}}};
    \node[draw, rectangle, minimum height=1.3cm](3) at (6,0) {\footnotesize{classifier}};
    \node(4) at (8,0) {\footnotesize{decision}};
    \draw[->,>=stealth']
    (0) edge (1)
    (1) edge (2)
    (2) edge (3)
    (3) edge (4);
    \end{tikzpicture}
    \caption{Architecture of a model based on DNN.}
    \label{fig:workflow}
\end{figure}
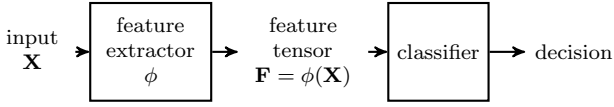

In this work we consider a deep learning framework to address classification problems. Such a framework can be interpreted, as depicted in Figure~\ref{fig:workflow}, as the concatenation of: 1) a feature extractor $\phi$ which is used to map inputs of raw images (matrix $\mathbf{X}\in\mathbb{R}^{n\times l}$, where $n$ is the number of samples and $l$ their dimensions) into high-level easily exploitable features (matrix $\mathbf{F}\in\mathbb{R}^{n\times d}$, where $d$ is the dimension of the feature vectors), and 2) a classifier meant to treat feature vectors as inputs. The purpose of the proposed methodology is to insert an intra-class denoising procedure between these two steps \emph{after the training of the feature extractor}. Once denoised feature vectors have been obtained, only the classifier is retrained for improved accuracy. In the case of few-shot learning, the feature extractor is typically trained on a generic dataset different from the one used to train the classifier~\cite{hu2020leveraging}. Before detailing the proposed denoising solution, let us first introduce some necessary notions and tools.

\subsection{Graph signal processing and graph filters}

The framework of GSP allows to manipulate signals defined on graph structures~\cite{shuman2013emerging, ortega2018graph}. It is then possible to design tools analogous to classical Fourier analysis. We are in particular interested in graph filtering. To introduce these tools, we first need to define graphs. A graph $\mathcal{G} = \langle \mathcal{V}, \mathcal{E}\rangle$ is a tuple made of a countable set $\mathcal{V}$, called vertices, and a set of pairs of vertices $\mathcal{E}$, called edges. Such a graph is usually represented by its binary ($\{0,1\}$) symmetric adjacency matrix $\mathbf{W}$ defined as $\mathbf{W}_{i,j} = 1$ if and only if $(i,j)\in \mathcal{E}$. This matrix can be extended to represent weighted graphs ($\mathbf{W}_{i,j}\in\mathbb{R}^+$).

The diagonal degree matrix $\mathbf{D}$ of $\mathcal{G}$ is defined as:
\begin{equation}
\mathbf{D} = \mathrm{diag}\left( \left[\sum_{j=1}^{|\mathcal{V}|} \mathbf{W}_{i,j}\right]_{i \in \{1, \dots, |\mathcal{V}|\}} \right)\;.
\end{equation}

If all vertices in $\mathcal{G}$ are connected to at least one other vertex, it is possible to define the normalized Laplacian of the graph as the matrix $\mathbf{L}$:
\begin{equation}
\mathbf{L} = \mathbf{I} - \mathbf{D}^{-\frac{1}{2}} \mathbf{W} \mathbf{D}^{-\frac{1}{2}}\;.
\end{equation}

Being symmetric and real-valued, $\mathbf{L}$ can be decomposed into a matrix of orthogonal eigenvectors $\mathbf{U}$, and corresponding nonnegative eigenvalues $\boldsymbol{\lambda}$, with $\lambda_1 \leq \dots \leq \lambda_{|\mathcal{V}|}$, such as $\mathbf{L} = \mathbf{U} \mathrm{diag}(\boldsymbol{\lambda}) \mathbf{U}^\top$.

In the field of GSP, we are interested in manipulating signals over $\mathcal{G}$. A signal is a vector $\mathbf{x}\in \mathbb{R}^{|\mathcal{V}|}$. The product $\hat{\mathbf{x}} = \mathbf{U}^\top \mathbf{x}$ is called the Graph Fourier Transform (GFT) of $\mathbf{x}$. The inverse Graph Fourier Transform (iGFT) is then defined as $\mathbf{x} = \mathbf{U} \hat{\mathbf{x}}$.

By analogy with classical Fourier analysis, values in $\boldsymbol{\lambda}$ are interpreted as frequencies and can thus be used to filter a signal. As such, a filter is usually defined by its spectral response: $h: \lambda \mapsto h(\lambda)$, where $\lambda \in \boldsymbol{\lambda}$. By writing $\mathbf{H} = \mathrm{diag}(h(\boldsymbol{\lambda}))$, the filtered signal $\mathbf{x}^{\mathrm{filter}}$ is then defined as:
\begin{equation}
\mathbf{x}^{\mathrm{filter}} = \mathbf{U}\mathbf{H}\mathbf{U}^\top \mathbf{x}\;.
\end{equation}

To remove high frequencies, a typical low-pass filter $h$ would nullify large values of $\lambda$.

\subsection{Proposed methodology}
\label{sec:methodo}

In this work we propose as stated before to filter out the high frequencies within the feature representation of data to improve classification performance. To achieve this, we first infer graphs from labeled feature vectors within each class and then remove high frequencies using low-pass graph filters.

More precisely, we proceed as follows. Consider all labeled signals in $\mathcal{D}_{\mathrm{base}}$ of class $c$. We denote these signals as $\mathbf{X}$, and the associated features $\mathbf{F} = \phi(\mathbf{X})$ (see Figure \ref{fig:workflow}). We first define a similarity matrix $\mathbf{S}$ between samples in the feature space as follows: 
\begin{equation}
 \mathbf{S}_{i,j}= s\left(\mathbf{F}_{i,:}, \mathbf{F}_{j,:}\right)\text{~if~} i \neq j \text{~and~} 0 \text{~otherwise}\;,
\end{equation}
where $s$ is a similarity measure. In our work, we choose cosine similarity for $s$.

Given a similarity matrix $\mathbf{S}$, we generate the adjacency matrix $\mathbf{W}$ of a graph $\mathcal{G}$ for each class using $k$-nearest neighbors selection: $\mathbf{W}_{i,j} = \mathbf{S}_{i,j}$ if $\mathbf{S}_{i,j}$ is among the $k$ largest entries of $\mathbf{S}_{i,:}$ or $\mathbf{S}_{:,j}$, and 0 otherwise.

The obtained graph is then used to define a low-pass filter $\mathbf{H}$ that we apply on $\mathbf{F}$ as follows:
\begin{equation}
\mathbf{F}^{\mathrm{filter}} = \mathbf{U}\mathbf{H}\mathbf{U}^\top \mathbf{F} \;,
\end{equation}
where $\mathbf{U}$ are the eigenvectors of the Laplacian matrix built from $\mathbf{W}$. Finally, we substitute $\mathbf{F}$ with $\mathbf{F}^{\mathrm{filter}}$ as an input for the classifier in the workflow in Figure \ref{fig:workflow}.

\section{Effect of Low-pass Graph Filters on Centroids}
\label{sec:theory} 

We would like to show that graph filtering can have the benefits of both keeping the expectation of centroids of feature vectors invariant while reducing their covariance. As such, graph filtering has the effect of reducing intra-class noise when training the classifier. To this end, we consider a simplified case where data is drawn from a Gaussian model. This is not an undesirable model since many works hypothesize that the features from the same class are aligned with a such a specific distribution~\cite{hu2020leveraging}.

Let $\mathbf{X} \in \mathbb{R}^{m\times l}$ be the subset of labeled samples of class $c$ in $\mathcal{D}_{\mathrm{base}}$, and $\mathbf{F} = \phi(\mathbf{X})$ the associated features ($\mathbf{F} \in \mathbb{R}^{m\times d}$, see Figure \ref{fig:workflow}). In the remainder of this section, we are interested in the distribution of the centroid $\boldsymbol{\gamma} \in \mathbb{R}^{d}$ obtained from features in $\mathbf{F}$. It is defined as:
\begin{equation}
\boldsymbol{\gamma} = \dfrac{1}{m} \sum_{i=1}^{m} \mathbf{F}_{i,:} \;.
\end{equation}

In the following lemma we provide analytical expressions for the mean and the covariance of the filtered centroid, and show their relations with those of the centroid of the original feature vectors.

\begin{lemma}
\label{meancov}
For all $i$, suppose $\mathbf{F}_{i,:}$ are i.i.d such that $\mathbf{F}_{i,:} \sim \mathcal{N}(\boldsymbol{\mu}, \sigma^2 \mathbf{I})$. Denote $\mathbf{L}$ the Laplacian of a graph obtained from $\mathbf{F}$ as described in Section \ref{sec:methodo}, with eigenvectors $\mathbf{U}$ and eigenvalues $\boldsymbol{\lambda}$. Choose $k \in \{1, \dots, m \}$, and define filter $\mathbf{H}$ such that $h(\lambda)=1$ if $\lambda \leq \lambda_k$ and $h(\lambda)=0$ otherwise ($\lambda \in \boldsymbol{\lambda}$). The mean and the covariance of the filtered centroid $\boldsymbol{\gamma}^{\mathrm{filter}}$ are given by:
\begin{equation}
\mathbb{E}\left(\boldsymbol{\gamma}^{\mathrm{filter}}\right) = \dfrac{1}{m} \sum_{j=1}^{k} \left(\mathbf{1}_m^\top\mathbf{U}_{:,j}\right)^2 \mathbb{E}\left(\boldsymbol{\gamma}\right) \;,
\end{equation}
\begin{equation}
 \mathrm{Cov}\left(\boldsymbol{\gamma}^{\mathrm{filter}}\right)=\dfrac{1}{m}\left(\sum_{i=1}^{m} \left(\mathbf{U}_{:, :k} \mathbf{U}_{:, :k}^\top\right)_{i,:} \right)^2 \mathrm{Cov}\left(\boldsymbol{\gamma}\right) \;,
\end{equation}
where $\mathbf{1}_m$ is the all-one column vector of dimension $m$.

\end{lemma}

\begin{proof}
See Section \ref{appendix}.
\end{proof}

Notice that the mean and the covariance of the filtered centroid $\boldsymbol{\gamma}^{\mathrm{filter}}$ obtained in Lemma \ref{meancov} are a weighted version of the mean and the covariance of the original centroid $\boldsymbol{\gamma}$, respectively. In the following Corollary \ref{cor:1}, under some conditions on the chosen graph and eigenvectors, we quantify those weights and discuss their effects on the centroids. We show that $\boldsymbol{\gamma}^{\mathrm{filter}}$ exhibits a lower covariance compared to $\boldsymbol{\gamma}$ while keeping the same expectation asymptotically.
\begin{corollary}
\label{cor:1}
Assume that we build a complete graph (i.e., a simple undirected graph in which every pair of distinct vertices is connected by a unique edge) for each class, and under the assumption of Lemma \ref{meancov}, we have:
\begin{equation}
    \mathbb{E}(\boldsymbol{\gamma}^{\mathrm{filter}}) = \dfrac{1}{(1-\frac{1}{m})} \mathbb{E}(\boldsymbol{\gamma})\;,
\end{equation}
\begin{equation}
    \mathrm{Cov}\left(\boldsymbol{\gamma}^{\mathrm{filter}}\right)  = \dfrac{1}{m(1-\frac{1}{m})^2}\mathrm{Cov}(\boldsymbol{\gamma})\;.
\end{equation}

In particular, by letting the number of labeled samples of that class $m \rightarrow \infty$, we get:
\begin{equation}
\mathbb{E}(\boldsymbol{\gamma}^{\mathrm{filter}}) = \mathbb{E}(\boldsymbol{\gamma}) \;,
\end{equation}
\begin{equation}
\mathrm{Cov}(\boldsymbol{\gamma}^{\mathrm{filter}}) = o
\left( \mathrm{Cov}(\boldsymbol{\gamma}) \right)\;.
\end{equation}

\end{corollary}

\begin{proof}
See Section \ref{appendix}.
\end{proof}

While this result holds only for the case of i.i.d feature vectors and a specific choice of a graph and a filter, we conjecture that it provides us with interesting insights for other cases as well, as shown experimentally. More specifically, the numerical results in the next section show that the proposed method improves the accuracy on real datasets and competitive classifiers.

\section{Numerical Results}
\label{sec:numerical-result}
In this section we follow the framework described in Figure \ref{fig:workflow} to conduct our numerical analysis. We show that graph filtering improves the performance of different classification methods in the few-shot and standard settings. The filter we use for our experiments is defined as:
\begin{align}
\label{equ:1}
\mathbf{H}_{i,i} & = \begin{cases} 1 & \text{if~}  \ i \leq k_1 \;,\\
 0.6 & \text{if~} k_1 \leq i \leq k_2 \;,\\
0 &  \text{otherwise}\;, \\
 \end{cases}
\end{align}
where $(k_1, k_2)  \in \{1, \dots, m\}$ are fixed for each scenario. In our experiments we came with empirically chosen values for $k_1$ and $k_2$, as well as for the choice of the value 0.6. They are not necessarily optimal but gave consistent results across datasets. Finding the best design of a graph filter is a direction of research for future work.

\subsection{Few-shot classification}
In the few-shot scenario we first use a pre-trained DNN (backbone) on a bigger dataset of base classes. We then use the backbone to perform transfer learning to the novel classes. For our experiments and following the standard procedure in the field, at each test run $5$ classes are drawn uniformly at random among the novel classes. For each class, $m=5$ labeled samples and $q=15$ unlabeled samples are uniformly drawn at random. We perform 100,000 iterations and report the mean accuracy and 95\% confidence intervals for each test.
\subsubsection{ Feature extractors and Datasets}
We consider two feature extractors. The
first one is a \textbf{Wide Residual Network} denoted \textbf{WRN} \cite{zagoruyko2016wide} and \textbf{Dense Networks} denoted \textbf{DNet121} \cite{wang2019simpleshot}\footnote{\url{https://github.com/yhu01/PT-MAP}.}.

We perform our experiments on four benchmark datasets:  \textbf{MiniImageNet} \cite{vinyals2016matching} denoted \textbf{MINet}, \textbf{CUB} \cite{wahcaltech}, \textbf{CIFAR-FS} \cite{bertinetto2018meta} and \textbf{TieredImageNet} \cite{ren2018meta} denoted \textbf{TINet}. These datasets are split into two parts: the base classes that are chosen to train the feature extractor and novel classes.

\subsubsection{Results}
 
To reduce noise on the labeled features, at each iteration we form $5$ graphs, each of $5$ nodes corresponding to the labeled features in each class. Then, we apply on them the low-pass filter described in (\ref{equ:1}) \cite{ortega2018graph}.  In the few-shot scenario we have graphs with 5 nodes and we fix the value of $(k_1, k_2)$ to $(1,4)$.  The results in Table \ref{genneral-accuracy} show that the method brings gains on the  performance of state-of-the-art classification method \cite{hu2020leveraging} in the case of $5$-shot, for both \textbf{WRN} and \textbf{DNet121} backbones and for all the considered datasets.

\begin{table}[ht]
\small
\centering
\begin{tabular}{|c|c|c|c|}
 \hline 
  \multicolumn{4}{|c|}{ \textbf{PT-MAP} \cite{hu2020leveraging}}  \\
  \hline
 Dataset  & Backbone & No Filter \%  & With Filter \%  \\ 
 \hline
 \multirow{2}{*}{ \textbf{MINet}} &  \textbf{WRN}  & 88.82  $\pm$ 0.013 & $\mathbf{88.90 } \pm \mathbf{0.011}$   \\
   &  \textbf{DNet121}  & $86.82  \pm 0.014$ & $\mathbf{86.87 } \pm \mathbf{0.014}$  \\ 
  \hline
   \textbf{CUB}  &  \textbf{WRN}  & 93.99  $\pm$ 0.011 & $\mathbf{94.14 } \pm \mathbf{0.009}$ \\ 
   \hline
 \textbf{ CIFAR} &  \textbf{WRN}  & 90.68  $\pm$ 0.015 &  $\mathbf{90.75} \pm \mathbf{0.015}$  \\ 
   \hline
 \textbf{TINet} &  \textbf{DNet121}  & $90.44\pm 0.014$  & $\mathbf{90.62  } \pm \mathbf{0.013}$ \\ 
   \hline
\end{tabular}
\caption{5-shot accuracy of the proposed methodology compared with state-of-the-art method in the literature, for various backbones. 
}
\label{genneral-accuracy}
\end{table}

\begin{figure}[h!]
\begin{center}
\begin{tikzpicture}
		\begin{axis}[	grid= major ,
				width=0.40\textwidth ,
				xlabel = {$m$-shot} ,
				ylabel = {Accuracy} ,
				xmin = 1, xmax = 5,
				ymin = 91.45, ymax = 94.40,
				height = 5cm,
				legend entries={ Without Filter, With Filter},
				legend style={at={(1,0)},anchor=south east}]
			\addplot coordinates {(1,91.55) (2,93.37) (3,93.77) (4,93.98) (5,93.99) }; 
			\addplot coordinates {(1, 91.55) (2,93.40) (3,93.82) (4,94.05) (5,94.14) }; 
		\end{axis}
	\end{tikzpicture}
	\end{center}
	\vspace{-.7cm}
	\caption{ Evolution of the accuracy on CUB (backbone: WRN) as a function of $m$.}
	\label{fig:accuracy-n}
	\end{figure}
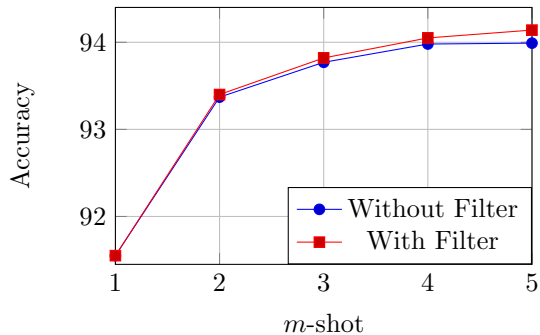

\begin{table}[ht]
\vspace{-.3cm}
\small
\centering
\begin{tabular}{|l|c|c|c|}
 \hline 
   & \multicolumn{3}{c|}{ \textbf{CIFAR-10}}  \\
   \hline
    Method   & \textbf{WRN} & \textbf{ShakeNet}  & \textbf{PyramidNet} \\
    \hline
    original paper & 95.82 \% &97.96 \% & 98.56 \% \\
NCM  & 85.81 \%  & \textbf{97.97} \% & 98.52 \% \\ 
   
1-NN  & 95.81 \% &  97.95 \% & 98.54 \% \\ 
 
   \hline
1-NN+Filter & \textbf{95.92} \% & \textbf{97.97} \% & \textbf{98.61} \%  \\
   \hline
\end{tabular}
\caption{Accuracy on the \textbf{CIFAR-10} dataset.  
}
\label{accuracy-cifar}
\end{table}
The total number of labeled samples $m$ has an impact on the classification performance when using graph filters. To verify that, we evaluate our proposed method while varying $m$. The results of the experiments are presented in Figure \ref{fig:accuracy-n}. We notice that the more we increase $m$ in each class, the better the performance. As a consequence, we observe the same results as those expected after our simplified analysis in Corollary \ref{cor:1}, which experimentally supports our conjecture.
\vspace{0.3cm}

\vspace{-.7cm}

\subsection{Standard classification}
In the standard classification scenario we use the well known \textbf{CIFAR-10} dataset and three pre-trained architectures, \textbf{WRN} \cite{zagoruyko2016wide}, \textbf{ShakeNet} \cite{gastaldi2017shake} and \textbf{PyramidNet} \cite{han2017deep}. The first model is trained with traditional data augmentation techniques (namely random crop and horizontal flip) while the latter two use a stronger learned policy called fast-autoaugment \cite{lim2019fast}. We extract the features, create $k$-nearest neighbor graphs for each class $(k=10)$. We apply the graph filter defined in (\ref{equ:1}) on each graph and generate our filtered features. Here, graphs have 5000 nodes. Parameters $(k_1, k_2)$ are set to (20, 55). Then, we compare the performance of a 1-NN classifier applied to the filtered features with the performance on the original datasets. The obtained results are described in Table \ref{accuracy-cifar}, where 1\% accuracy corresponds to 100 well-classified test images. The 1-NN classifier on the filtered features was able to improve the performance over both the 1-NN classifier and nearest class mean classifier (NCM) applied without denoising, and even beats the performance of untouched full DNN architecture~\cite{mangla2020charting}.
\section{ Conclusion}
\label{sec:conclusion}

We proposed a graph-based method to improve the accuracy of classification methods. The method consists in using techniques from GSP to reduce the noise in the extracted feature vectors that may affect the networks performance. We showed the effectiveness of the method theoretically. In addition, we performed experiments on standardized vision datasets and we obtained gains in two different settings: few-shot classification and standard classification. A possible future work would be to have an automatic way of choosing the best filter parameters or even integrating them as parameters during the learning phase. 

%
%
\bibliographystyle{IEEEtran}
\bibliography{references}

\begin{thebibliography}{10}
\providecommand{\url}[1]{#1}
\csname url@samestyle\endcsname
\providecommand{\newblock}{\relax}
\providecommand{\bibinfo}[2]{#2}
\providecommand{\BIBentrySTDinterwordspacing}{\spaceskip=0pt\relax}
\providecommand{\BIBentryALTinterwordstretchfactor}{4}
\providecommand{\BIBentryALTinterwordspacing}{\spaceskip=\fontdimen2\font plus
\BIBentryALTinterwordstretchfactor\fontdimen3\font minus
  \fontdimen4\font\relax}
\providecommand{\BIBforeignlanguage}[2]{{%
\expandafter\ifx\csname l@#1\endcsname\relax
\typeout{** WARNING: IEEEtran.bst: No hyphenation pattern has been}%
\typeout{** loaded for the language `#1'. Using the pattern for}%
\typeout{** the default language instead.}%
\else
\language=\csname l@#1\endcsname
\fi
#2}}
\providecommand{\BIBdecl}{\relax}
\BIBdecl

\bibitem{zhao2019object}
Z.-Q. Zhao, P.~Zheng, S.-t. Xu, and X.~Wu, ``Object detection with deep
  learning: A review,'' \emph{IEEE transactions on neural networks and learning
  systems}, vol.~30, no.~11, pp. 3212--3232, 2019.

\bibitem{lu2007survey}
D.~Lu and Q.~Weng, ``A survey of image classification methods and techniques
  for improving classification performance,'' \emph{International journal of
  Remote sensing}, vol.~28, no.~5, pp. 823--870, 2007.

\bibitem{shuman2013emerging}
D.~I. Shuman, S.~K. Narang, P.~Frossard, A.~Ortega, and P.~Vandergheynst, ``The
  emerging field of signal processing on graphs: Extending high-dimensional
  data analysis to networks and other irregular domains,'' \emph{IEEE signal
  processing magazine}, vol.~30, no.~3, pp. 83--98, 2013.

\bibitem{torrey2010transfer}
L.~Torrey and J.~Shavlik, ``Transfer learning,'' in \emph{Handbook of research
  on machine learning applications and trends: algorithms, methods, and
  techniques}.\hskip 1em plus 0.5em minus 0.4em\relax IGI Global, 2010, pp.
  242--264.

\bibitem{rohrbach2013transfer}
M.~Rohrbach, S.~Ebert, and B.~Schiele, ``Transfer learning in a transductive
  setting,'' in \emph{Advances in neural information processing systems}, 2013,
  pp. 46--54.

\bibitem{hu2020leveraging}
Y.~Hu, V.~Gripon, and S.~Pateux, ``Leveraging the feature distribution in
  transfer-based few-shot learning,'' \emph{arXiv preprint arXiv:2006.03806},
  2020.

\bibitem{ortega2018graph}
A.~Ortega, P.~Frossard, J.~Kova{\v{c}}evi{\'c}, J.~M. Moura, and
  P.~Vandergheynst, ``Graph signal processing: Overview, challenges, and
  applications,'' \emph{Proceedings of the IEEE}, vol. 106, no.~5, pp.
  808--828, 2018.

\bibitem{zagoruyko2016wide}
S.~Zagoruyko and N.~Komodakis, ``Wide residual networks,'' \emph{arXiv preprint
  arXiv:1605.07146}, 2016.

\bibitem{wang2019simpleshot}
Y.~Wang, W.-L. Chao, K.~Q. Weinberger, and L.~van~der Maaten, ``Simpleshot:
  Revisiting nearest-neighbor classification for few-shot learning,''
  \emph{arXiv preprint arXiv:1911.04623}, 2019.

\bibitem{vinyals2016matching}
O.~Vinyals, C.~Blundell, T.~Lillicrap, D.~Wierstra \emph{et~al.}, ``Matching
  networks for one shot learning,'' in \emph{Advances in neural information
  processing systems}, 2016, pp. 3630--3638.

\bibitem{wahcaltech}
C.~Wah, S.~Branson, P.~Welinder, P.~Perona, and S.~Belongie, ``The caltech-ucsd
  birds-200-2011 dataset (2011),'' \emph{California Institute of Technology}.

\bibitem{bertinetto2018meta}
L.~Bertinetto, J.~F. Henriques, P.~H. Torr, and A.~Vedaldi, ``Meta-learning
  with differentiable closed-form solvers,'' \emph{arXiv preprint
  arXiv:1805.08136}, 2018.

\bibitem{ren2018meta}
M.~Ren, E.~Triantafillou, S.~Ravi, J.~Snell, K.~Swersky, J.~B. Tenenbaum,
  H.~Larochelle, and R.~S. Zemel, ``Meta-learning for semi-supervised few-shot
  classification,'' \emph{arXiv preprint arXiv:1803.00676}, 2018.

\bibitem{gastaldi2017shake}
X.~Gastaldi, ``Shake-shake regularization,'' \emph{arXiv preprint
  arXiv:1705.07485}, 2017.

\bibitem{han2017deep}
D.~Han, J.~Kim, and J.~Kim, ``Deep pyramidal residual networks,'' in
  \emph{Proceedings of the IEEE conference on computer vision and pattern
  recognition}, 2017, pp. 5927--5935.

\bibitem{lim2019fast}
S.~Lim, I.~Kim, T.~Kim, C.~Kim, and S.~Kim, ``Fast autoaugment,'' in
  \emph{Advances in Neural Information Processing Systems}, 2019, pp.
  6665--6675.

\bibitem{mangla2020charting}
P.~Mangla, N.~Kumari, A.~Sinha, M.~Singh, B.~Krishnamurthy, and V.~N.
  Balasubramanian, ``Charting the right manifold: Manifold mixup for few-shot
  learning,'' in \emph{The IEEE Winter Conference on Applications of Computer
  Vision}, 2020, pp. 2218--2227.

\end{thebibliography}

\section{appendices}
\label{appendix}

\section*{Proof of Lemma \ref{meancov}}

In this Appendix, we provide analytical expressions for the mean and the covariance of the filtered centroids as a function of the mean and covariance of the original centroids.
By applying the chosen filter $\mathbf{H}$ to $\mathbf{F}$, we get:
\begin{align}
\boldsymbol{\gamma}^{\mathrm{filter}} & = \dfrac{1}{m} \sum_{i=1}^{m} \left(\mathbf{U} \mathbf{H} \mathbf{U}^\top \mathbf{F}\right)_{i,:} \\
& = \dfrac{1}{m} \sum_{i=1}^{m} \left(\mathbf{U}_{, :k} \mathbf{U}_{:, :k}^\top \mathbf{F}\right)_{i,:}\;.
\end{align}

Let us compute the expectancy of $\boldsymbol{\gamma}^{\mathrm{filter}}$:
\begin{align}
  \mathbb{E}\left(\boldsymbol{\gamma}^{\mathrm{filter}}\right) & =
  \mathbb{E}\left(\dfrac{1}{m} \sum_{i=1}^{m} \left(\mathbf{U}_{:, :k} \mathbf{U}_{:, :k}^\top \mathbf{F}\right)_{i,:}\right) \\
  & = \dfrac{1}{m} \mathbb{E}\left(\sum_{i=1}^{m} \left(\mathbf{U}_{:, :k} \mathbf{U}_{:, :k}^\top \mathbf{F}\right)_{i,:}\right) \\
  & \simeq \dfrac{1}{m} \mathbf{1}_m^\top \mathbf{U}_{:, :k} \mathbf{U}_{:, :k}^\top \mathbf{1}_m \mathbb{E}\left(\boldsymbol{\gamma}\right) \\
  & = \dfrac{1}{m} \sum_{j=1}^{k} \left(\mathbf{1}_m^\top\mathbf{U}_{:,j}\right)^2 \mathbb{E}\left(\boldsymbol{\gamma}\right).
\end{align}
The last step follows because $ \mathbf{1}_m^\top \mathbf{U}_{:, :k} \mathbf{U}_{:, :k}^\top \mathbf{1}_m$ can be linked to the eigenvectors of $\mathbf{U}_{:, :k}$ as follows: $ \mathbf{1}_m^\top \mathbf{U}_{:, :k} \mathbf{U}_{:, :k}^\top \mathbf{1}_m = (\mathbf{U}_{:, :k}^\top \mathbf{1}_m)^\top (\mathbf{U}_{:, :k}^\top \mathbf{1}_m ) = \sum_{j=1}^{k} (\mathbf{1}_m^\top\mathbf{U}_{:,j})^2,$ where $\mathbf{U}_{:,j}$ denotes the $j$-th unit eigenvector (i.e., the $j$-th column of $\mathbf{U}_{:, :k}$ and $\mathbf{1}_{m}^{\top}\mathbf{U}_{:,j}$ is the sum of the entries of $\mathbf{U}_{:,j}$. 

Now, let us study the covariance of $\boldsymbol{\gamma}^{\mathrm{filter}}$:
\begin{align}
 & \mathrm{Cov}\left(\boldsymbol{\gamma}^{\mathrm{filter}}\right)  = \mathrm{Cov}\left(\dfrac{1}{m} \sum_{i=1}^{m} \left(\mathbf{U}_{:, :k} \mathbf{U}_{:, :k}^\top \mathbf{F}\right)_{i,:}\right) \\
  & \overset{(a)}{\simeq} \dfrac{\sigma^2}{m^2} \mathrm{diag}\left(\left[\left(\sum_{i=1}^{m} \left(\mathbf{U}_{:, :k} \mathbf{U}_{:, :k}^\top\right)_{i,:} \right)^2 \right]_{j \in \{1, \dots, d\}}\right)
\end{align}


Steps $(a)$ follow from the i.i.d. Gaussian distribution assumption on the variables $\mathbf{F}_{i,:}$. 
 \qed

\section*{Proof of Corollary \ref{cor:1}}

In this Appendix, we quantify the weights we obtained in Lemma \ref{meancov}. We show that under some conditions $\boldsymbol{\gamma}^{\mathrm{filter}}$ exhibits a lower covariance compared to $\boldsymbol{\gamma}$ while keeping the same expectation asymptotically.

In the case of complete graphs, the degree of each vertex is $m-1$. Let $k=1$ such that $\mathbf{U}_{:, :k}$ will contain only the first eigenvector, i.e.:
\begin{equation}
    \mathbf{U}_{:, :k} = \textbf{U}_{:,1} = \left[\frac{1}{\sqrt{m-1}},...,\frac{1}{\sqrt{m-1}}\right]^\top \;.
\end{equation}

We therefore have:
\begin{equation}
    \mathbf{U}_{:, :k} \mathbf{U}_{:, :k}^\top = \frac{1}{m-1} \mathbf{1}_{m \times m}\;,
\end{equation}
where $\mathbf{1}_{m \times m}$ is the all-one matrix dimension $m \times m$.

For this special case, we can show that the original and filtered centroids have the same mean, and that covariance of the filtered centroids decreases as the number of the labeled samples $m$ grows, i.e.:
\begin{align}
\mathbb{E}(\boldsymbol{\gamma}^{\mathrm{filter}}) & = \dfrac{1}{m} \sum_{j=1}^{k} (\mathbf{1}_m^{\top}\textbf{U}_{:,j})^2 = \dfrac{1}{(1-\frac{1}{m})} \mathbb{E}(\boldsymbol{\gamma}) \;,
\end{align}
\begin{align}
\mathrm{Cov}\left(\boldsymbol{\gamma}^{\mathrm{filter}}\right) & = \dfrac{1}{m}\sum_{i=1}^{m} \left(\mathbf{U}_{:, :k} \mathbf{U}_{:, :k}^\top\right)_{i,:} \mathrm{Cov}\left(\boldsymbol{\gamma}\right) \\
& = \dfrac{1}{m(1-\frac{1}{m})^2}\mathrm{Cov}(\boldsymbol{\gamma}) \;.
\end{align}

When $m \rightarrow \infty$, the result follows.  \qed


\end{document}